\documentclass[12pt,a4paper]{article}

\usepackage[T1]{fontenc}
\usepackage[utf8]{inputenc}
\usepackage[english]{babel}
\usepackage{geometry}
\usepackage{authblk}
\usepackage{booktabs}
\usepackage{longtable}
\usepackage{array}
\usepackage{parskip}
\usepackage{setspace}
\usepackage{titlesec}
\usepackage{xcolor}
\usepackage{graphicx}
\usepackage{tikz}
\usepackage{float}
\usetikzlibrary{arrows.meta, positioning, shapes.geometric}
\usepackage{caption}
\usepackage[style=vancouver, sorting=none, maxbibnames=3, minbibnames=1, maxcitenames=3, mincitenames=1, backend=biber]{biblatex}
\AtEveryBibitem{\clearfield{url}}
\usepackage{textcase}

\usepackage{hyperref}
\hypersetup{colorlinks=true, linkcolor=blue, citecolor=blue, urlcolor=blue}

\titleformat{\section}{\large\bfseries}{\thesection.}{0.5em}{}
\titleformat{\subsection}{\normalsize\bfseries\itshape}{\thesubsection}{0.5em}{}
\titleformat{\subsubsection}{\normalsize\itshape}{\thesubsubsection}{0.5em}{}
\defbibheading{bibliography}[\bibname]{\section*{REFERENCES}}

\begin{document}

\begin{center}
  \parbox{1.0\textwidth}{%
    \centering\LARGE\bfseries\linespread{1.0}\selectfont
    EuropeMedQA Study Protocol: A Multilingual, Multimodal
    Medical Examination Dataset for Language Model Evaluation}
\end{center}

\vspace{0.8em}

\small\noindent
Francesco Andrea Causio\textsuperscript{1,2},
Vittorio De Vita\textsuperscript{1,2},
Olivia Riccomi\textsuperscript{2},
Michele Ferramola\textsuperscript{8},
Federico Felizzi\textsuperscript{2},
Alessandro Tosi\textsuperscript{2},
Antonio Cristiano\textsuperscript{1,2},
Lorenzo De Mori\textsuperscript{2,3},
Chiara Battipaglia\textsuperscript{2},
Melissa Sawaya\textsuperscript{2},
Luigi De Angelis\textsuperscript{2,4},
Marcello Di Pumpo\textsuperscript{1,2},
Alessandra Piscitelli\textsuperscript{2,9},
Pietro Eric Risuleo\textsuperscript{1,2},
Alessia Longo\textsuperscript{7},
Giulia Vojvodic\textsuperscript{2,5},
Mariapia Vassalli\textsuperscript{2,11},
Bianca Destro Castaniti\textsuperscript{2,10},
Nicol\`{o} Scarsi\textsuperscript{1,2,6},
Manuel Del Medico\textsuperscript{1,2}*

\normalsize
\vspace{0.6em}

\noindent\textbf{Affiliations}
\vspace{-0.8em}
\begin{footnotesize}
\begin{enumerate}\setlength{\itemsep}{-0.1pt}
  \item {Section of Hygiene, University Department of Life Sciences and Public Health, Universit\`a Cattolica del Sacro Cuore, Rome, Italy}
  \item {Societ\`a Italiana Intelligenza Artificiale in Medicina (SIIAM), Rome, Italy}
  \item {Department of Mental Health, Addiction Treatment Unit, ASL RM 4, Bracciano, Italy}
  \item {Department of Translational Research and New Technologies in Medicine and Surgery, University of Pisa, Pisa, Italy}
  \item {Universit\`a Cattolica del Sacro Cuore, Rome, Italy}
  \item {Department of Medicine and Surgery, University of Perugia}
  \item {Universit\'e Paris Cit\'e, Paris, France}
  \item {NSBproject, Mantova, Italy}
  \item {Department of Psychiatry, University of Campania ``Luigi Vanvitelli'', Naples, Italy}
  \item {Department of Diagnostic Imaging, Oncological Radiotherapy and Hematology, Universit\`a Cattolica del Sacro Cuore, Rome, Italy}
  \item {Department of Experimental Medicine, Section of Medical Pathophysiology, Food Science and Endocrinology, Sapienza University, Rome, Italy}
\end{enumerate}
\vspace{-1.4em}
\end{footnotesize}
\begin{footnotesize}
{*Corresponding author: manuel.delmedico01@icatt.it}
\end{footnotesize}
\vspace{1em}
\hrule
\vspace{1em}
\begin{abstract}
While Large Language Models (LLMs) have demonstrated high proficiency on English-centric medical examinations, their performance often declines when faced with non-English languages and multimodal diagnostic tasks. This study protocol describes the development of EuropeMedQA, the first comprehensive, multilingual, and multimodal medical examination dataset sourced from official regulatory exams in Italy, France, Spain, and Portugal. Following FAIR data principles and SPIRIT-AI guidelines, we describe a rigorous curation process and an automated translation pipeline for comparative analysis. We evaluate contemporary multimodal LLMs using a zero-shot, strictly constrained prompting strategy to assess cross-lingual transfer and visual reasoning. EuropeMedQA aims to provide a contamination-resistant benchmark that reflects the complexity of European clinical practices and fosters the development of more generalizable medical AI.
\end{abstract}
\section{INTRODUCTION}

The increasing interest in applying large language models to medicine is driven in part by their impressive performance on medical exam questions, with models such as Med-PaLM~2 and GPT-4 achieving passing scores on the United States Medical Licensing Examination \cite{alonso2024,chen2024}. However, these examinations may not fully capture the complexity of real patient--doctor interactions, the synthesis of diverse medical literature, or the nuanced clinical decision-making required in practice \cite{alwakeel2025,federico2025}. Furthermore, LLMs produce varying outcomes when evaluated on items from different countries and contexts, owing to disparities in disease prevalence, clinical guidelines, terminologies, and question formats across regions \cite{alonso2024,grzybowski2024polish,rosol2023}. For instance, while GPT-4 excels on USMLE-style questions, performance drops notably on non-English exams like Polish medical licensing tests or Spanish benchmarks, revealing biases toward English-centric training data and limited generalizability \cite{alonso2024,grzybowski2024polish}. The proliferation of foundation models, both open-source and proprietary, further complicates the selection of models best suited for specific clinical applications \cite{riccomi2025}. Although numerous studies have assessed model accuracy on established datasets like MedQA and MedMCQA, the widespread online availability of these resources poses a significant risk of training data contamination, artificially inflating performance metrics for LLMs \cite{alwakeel2025,riccomi2025,askell2024}. Synthetic question generation has been proposed to circumvent this issue, yet it often fails to replicate the real-world complexity, cultural nuances, and clinical appropriateness of official exams \cite{kell2024}.

To address these challenges---including data contamination, English bias, lack of multimodality, and insufficient diversity in exam contexts---we intend to develop the EuropeMedQA dataset. EuropeMedQA positions itself distinctly in the literature as the first comprehensive European-focused, multilingual, multimodal dataset from real regulatory exams, bridging gaps left by predominantly English or non-European benchmarks like USMLE-derived MedQA, Indian MedMCQA, or global efforts like MultiMedQA \cite{alwakeel2025,cheng2024}. While prior works highlight LLMs' strengths on text-only US exams, they underscore limitations in cross-lingual transfer, visual reasoning, and handling country-specific practices---issues EuropeMedQA directly tackles \cite{alonso2024,grzybowski2024history}.

This benchmarking analysis study aims to investigate:
\begin{enumerate}
  \item The distribution of medical question topics and their intrinsic characteristics across diverse European national contexts.
  \item The performance of contemporary multimodal LLMs in accurately answering questions from the EuropeMedQA dataset. Furthermore, the study will comparatively evaluate the performance of both text-based and multimodal models on this novel dataset, providing insights into the added value of multimodal capabilities for complex medical question answering.
\end{enumerate}

\section{MATERIALS AND METHODS}

The EuropeMedQA dataset will be developed in accordance with the FAIR data principles, ensuring transparency, reproducibility, and reusability in artificial intelligence research \cite{mugahid2024}. Consistent with the SPIRIT-AI guidelines \cite{rivera2020}, the dataset architecture, data provenance, and annotation procedures will be systematically documented to enhance transparency in model development and evaluation. Comprehensive metadata enabling full reproducibility of the dataset construction process will be made publicly available alongside the dataset (e.g., via a dedicated GitHub repository).

\subsection{Study Design}

The EuropeMedQA study is designed as a prospective dataset collection and validation project, enabling both supervised and semi-supervised model training. The primary aim is to create a robust dataset that includes diverse multimodal data types and languages, facilitating comprehensive model evaluation in medical knowledge-related questions and answering.

\subsection{Search Strategy and Eligibility Criteria}

We will search online databases and Google for medical licensing examinations and residency admission exams in Italy (Scuola di Specializzazione Medica, SSM), France (Examen Classant National, ECN), Spain (M\'edico Interno Residente), and Portugal (Prova Nacional de Acesso, PNA).

\subsection{Inclusion Criteria}

\begin{itemize}
  \item Questions and examinations issued by a central government institution (e.g., Ministry of Education or Ministry of Health) or by officially recognized regulatory authorities responsible for the evaluation of medical professionals upon entering the medical profession (medical licensing) or before enrolling in a medical residency (medical residency admission).
  \item Questions available in the original language and in unedited form.
  \item Text-based questions and image-based questions where the related images are available.
  \item Image-based questions where corresponding images can be retrieved from official sources (ministerial documents and residency preparation materials containing authentic test questions).
\end{itemize}

\subsection{Exclusion Criteria}

\begin{itemize}
  \item Questions issued from unofficial sources (e.g., schools preparing doctors for medical licensing examinations).
  \item Image-based questions where corresponding images are not retrievable from official sources or not available.
\end{itemize}

\begin{figure}[H]
  \centering
  \vspace{1em}
  \resizebox{0.75\textwidth}{!}{%
  \begin{tikzpicture}[
      box/.style     = {rectangle, rounded corners=4pt, draw=black, fill=black,
                        text=white, font=\large\bfseries,
                        text width=5.2cm, align=center,
                        minimum height=1.2cm, inner sep=8pt},
      compbox/.style = {rectangle, rounded corners=4pt, draw=black, fill=black,
                        text=white, font=\Large\bfseries,
                        text width=5.2cm, align=center,
                        minimum height=1.2cm, inner sep=8pt},
      arrow/.style   = {-{Stealth[length=7pt]}, very thick},
  ]
  \node at (0,  1.5) {};
  \node[box] at (0,  0.0) (search)   {Internet search for medical licensing and residency admission tests};
  \node[box] at (0, -3.2) (retrieve) {Retrieve original test};
  \node[box] at (0, -5.2) (extract)  {Data extraction};
  \node[box] at (0, -7.2) (revise)   {Extraction revision};
  \node[box] at (0, -9.2) (clean)    {Dataset cleaning};
  \node[box] at (0,-11.2) (process)  {Dataset processing};
  \node[box] at (-6.5,-14.2) (transl)  {English translation\\(GPT-4o mini),\\validated};
  \node[box] at ( 0,  -14.2) (final)   {Final dataset in\\original languages};
  \node[box] at ( 6.5,-14.2) (further) {Further analysis\\(additional subsets TBD)};
  \node[box] at (-6.5,-17.2) (llmen)  {LLM API evaluates\\each question\\(English dataset)};
  \node[box] at ( 0,  -17.2) (llmori) {LLM API evaluates\\each question\\(original languages)};
  \node[compbox] at (0,-20.2) (compare) {COMPARISON};
  \node at (0,-21.5) {};
  \draw[arrow] (search)   -- (retrieve);
  \draw[arrow] (retrieve) -- (extract);
  \draw[arrow] (extract)  -- (revise);
  \draw[arrow] (revise)   -- (clean);
  \draw[arrow] (clean)    -- (process);
  \draw[arrow] (process.south) -- ++(0,-0.8) -| (transl.north);
  \draw[arrow] (process.south) -- ++(0,-0.8)  -- (final.north);
  \draw[arrow] (process.south) -- ++(0,-0.8) -| (further.north);
  \draw[arrow] (transl)  -- (llmen);
  \draw[arrow] (final)   -- (llmori);
  \draw[arrow] (llmen.south)  -- ++(0,-0.8) -| (compare.north);
  \draw[arrow] (llmori.south) -- (compare.north);
  \end{tikzpicture}%
  }
  \vspace{1em}
  \begin{small}     
   \caption{Study flow diagram. After dataset processing, three parallel tracks are produced: an English-translated version, the original-language dataset, and any additional subsets. LLM evaluation is performed on both the English and original-language tracks, and results are compared.}
  \label{fig:flowdiagram}
  \end{small}
\end{figure}
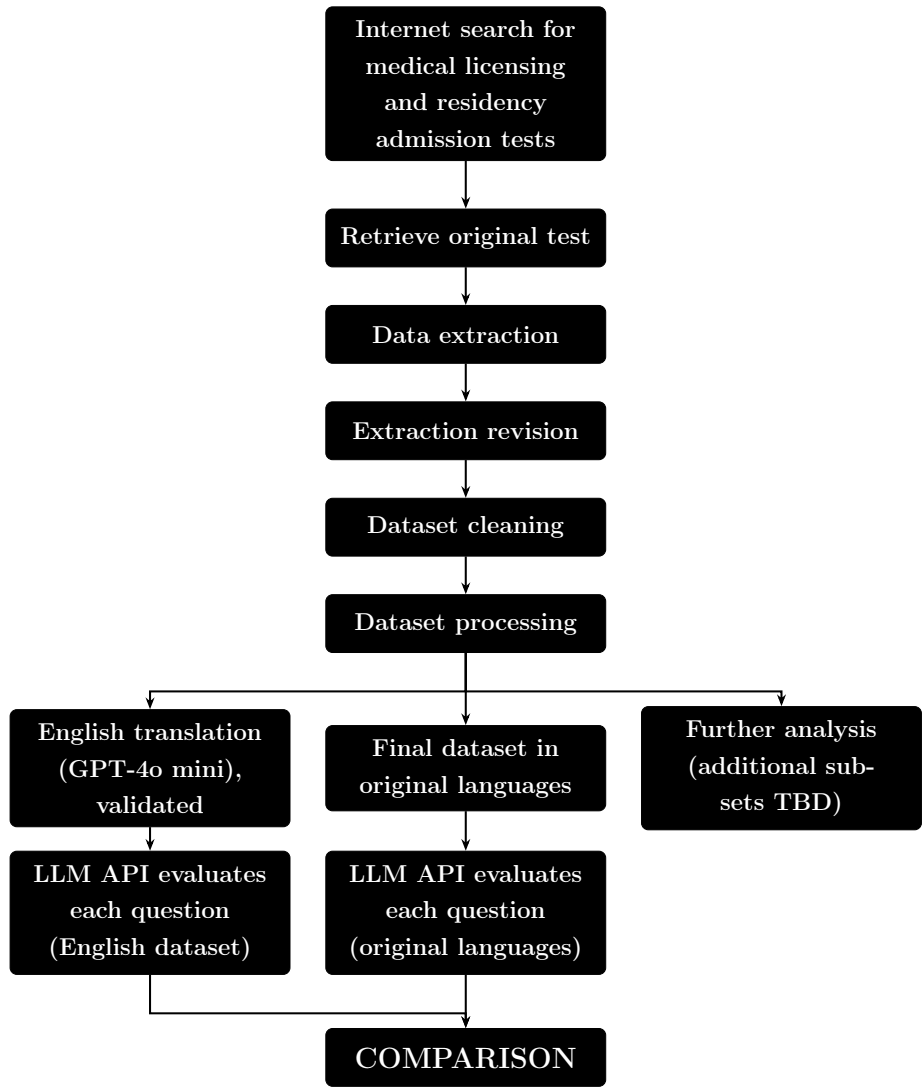

The study flow is summarized in Figure~\ref{fig:flowdiagram}.

\subsection{Data Collection and Extraction}

The dataset will consist of multimodal data, including structured information from various medical domains, ensuring multilingual diversity by sourcing content from medical texts in multiple European languages. Questions from the retrieved examinations will be extracted using a predefined and reproducible data extraction protocol supported by a standardized ChatGPT prompt. The extraction process will be performed by a team of six medical doctors (M.V., B.D.C., A.P., L.D.M., M.D.P., G.V), each fluent in at least one of the examination languages, to ensure linguistic and clinical accuracy. To enhance data validity and minimize extraction errors, all records will be independently cross-checked by six researchers (F.A.C., V.D.V., A.C., M.D.M., P.E.R., L.D.A.) with expertise in medical research methodology and dataset curation. Discrepancies will be resolved by consensus. All extracted variables will follow consistent naming conventions and predefined data types to ensure interoperability and facilitate downstream analyses. The data extraction framework and all collected variables are summarized in \textbf{Table~\ref{tab:extraction}}.

Images corresponding to each question were systematically identified through targeted web searches of official ministerial documents and medical residency preparation materials featuring authentic test questions and images. These were manually extracted and saved in high-quality, machine-readable \texttt{.png} format to prevent technical issues and minimize performance bias or distortion. For questions embedded in comprehensive PDF files with all associated test images, screenshots were utilized to capture and extract images, ensuring accurate textual correspondence with their respective questions and eliminating mismatches.

\vspace{0.5em}
    \noindent \begin{small}
        {Table~\ref{tab:extraction}.} Extraction Framework and Dataset Categories.\end{small}

\begin{longtable}{>{\ttfamily}p{0.28\textwidth} p{0.65\textwidth}}
  \toprule
  \normalfont\textbf{Variable} & \textbf{Description} \\
  \midrule
  \endfirsthead
  \toprule
  \normalfont\textbf{Variable} & \textbf{Description} \\
  \midrule
  \endhead
  \bottomrule
  \endlastfoot
  questionID        & A unique identifier for each question number in the dataset: country ID (IT/ES/FR/PT), as a progressive number. \\[4pt]
  testyear          & The year in which the test was administered to the public. \\[4pt]
  numberintest      & The position or number of the question within the test. \\[4pt]
  picture\_provided & A flag (yes/no) indicating whether a visual element (e.g., image) is based on the presence of an image (e.g., EKG). \\[4pt]
  question\_text    & The text of the question itself including a scenario description, when appropriate. \\[4pt]
  picture\_link     & A URL or reference to the shared Drive folder where the question picture is archived, if available. \\[4pt]
  option\_a         & Option A full-text. \\[4pt]
  option\_b         & Option B full-text. \\[4pt]
  option\_c         & Option C full-text. \\[4pt]
  option\_d         & Option D full-text. \\[4pt]
  option\_e         & Option E full-text. \\[4pt]
  correct\_option   & The complete text of the correct answer. \\[4pt]
  correct\_fulltext & Correct option full text (should be the same). 
  \label{tab:extraction}
\end{longtable}

The French medical specialization access test consists of a main clinical case, optionally accompanied by an image, to which multiple related questions refer that may themselves include images. To ensure consistency with the Italian, Spanish, and Portuguese datasets, the \texttt{question\_text} field for each question also includes the full text of the reference clinical case along with its image (if present). The specific image for each question, if any, is instead referenced via a link in the \texttt{picture\_link} field.

\subsection{Dataset Curation}

After extraction, all collected questions will undergo a rigorous curation and quality assurance process conducted by the research team. This process will be designed to minimize noise and bias while ensuring standardization across heterogeneous data sources. Specifically, the dataset will undergo several processing steps, including:

\begin{itemize}
  \item \textbf{Dataset cleaning:}
  \begin{itemize}
    \item Verify that the questions and answers in the original document match those on the worksheet;
    \item Where an image is provided, verify that it is correctly associated with the relevant question;
    \item Where no image is provided, mark NO and N/A in the appropriate columns;
    \item Ensure that there are no grammatical errors, punctuation errors, or typos;
    \item Standardise all ``YES'' and ``NO'' answers;
    \item Standardise all answers A, B, C, D, and E in capital letters;
    \item When there is a context common to several questions, it will be repeated for each question referring to it, placing it in square brackets, followed by the specific question;
    \item Standardise the font, spacing, and text positioning to provide greater order.
  \end{itemize}
  \item Duplicate questions identified across different sources will be removed.
  \item Translating and verifying non-English questions to compare LLMs' accuracy in both the questions' original language and English translation. Questions will be translated using ChatGPT~4o mini.
\end{itemize}

The final dataset will include:
\begin{itemize}
  \item The original medical residency admission tests and medical licensing examinations for each of the available countries;
  \item The English translation of the included questions;
  \item The images provided with the questions.
\end{itemize}

The French subset will contain items for which multiple answers may be simultaneously correct, reflecting the rules in the French ECN. Questions in the France subset of EuropeMedQA may be associated with up to two distinct images, reflecting the hierarchical structure of the clinical prompt.

This structured approach aims to support transparency, reproducibility, and methodological rigor in downstream model evaluation, in line with established reporting standards for AI research in healthcare \cite{sounderajah2025}.

\subsection{Evaluation Methodology and Setup: LLM Testing}

\subsubsection{Overview of the Evaluation Pipeline}

This section describes the complete evaluation pipeline that will be adopted to assess the performance of large language models on the EuropeMedQA benchmark. The pipeline is organized into a sequence of clearly defined stages, designed to ensure methodological transparency, comparability across models, and reproducibility of results. Specifically, we describe:

\begin{enumerate}
  \item The \textit{preprocessing steps} applied to the dataset prior to inference, including answer-option shuffling and dataset unification;
  \item The \textit{translation pipeline} used to generate non-English versions of the benchmark;
  \item The \textit{randomization strategy} adopted to mitigate positional bias;
  \item The model inference setup, including prompting constraints and handling of multimodal inputs;
  \item Dataset-specific evaluation rules;
  \item The evaluation metrics and reproducibility considerations.
\end{enumerate}

Each stage is implemented as a fully scripted and procedurally controlled procedure, and all models are evaluated on identical dataset configurations to ensure a fair and controlled comparison.

\subsubsection{Dataset Preparation}

Before proceeding with model evaluation, an initial exploratory data analysis will be conducted to verify that the dataset's characteristics align with the intended objectives. Since the goal was not to assess model robustness to class imbalance, but rather to compare models under idealized, bias-free conditions, we randomly redistributed correct answer positions.

Where class imbalance and positional bias are present, models may exploit construct-irrelevant shortcuts to achieve deceptively high accuracy, such as systematically selecting the majority class \cite{alonso2024}. To remove this spurious advantage and ensure a fair zero-shot evaluation with large language models, a standard multiple-choice test design strategy will be applied: the position of the correct option was permuted for each item, and the corresponding answer key was updated \cite{alwakeel2025}. This procedure reduces sensitivity to answer-option ordering \cite{chen2024} and aligns the dataset structure with established principles of multiple-choice assessment design. This step is particularly necessary for subsets in which correct answers were originally mapped to a fixed option label, which would otherwise introduce a strong positional bias.

Answer permutation will be implemented as a dedicated preprocessing step using a fixed pseudo-random seed (42), ensuring reproducibility \cite{chen2024,cheng2024} across runs and auditability of the evaluation pipeline. The resulting distribution of correct answers is expected to be approximately uniform across the available options, mirroring the structure of well-designed multiple-choice examinations.

To enable joint evaluation across multiple resources in the same language, datasets will be unified by assigning a unique identifier that preserves the original QuestionID while allowing aggregation across sources. This approach ensures traceability of item provenance without altering item content.

All items lacking a validated ground-truth label will be removed prior to inference, as they are not informative for performance estimation. These preprocessing steps ensure that subsequent analyses rely exclusively on validated items. While these preprocessing steps mitigate major sources of positional and class-distribution bias, additional residual limitations related to dataset composition and evaluation settings are discussed at the end of this section.

\subsubsection{Translation Pipeline}

Translation will be performed exclusively as a preprocessing step and never during model inference or evaluation. Translation will be conducted independently in four separate runs. In all cases, the original QuestionID will be preserved unchanged to maintain a one-to-one correspondence between original and translated items. Translation will be performed using GPT-4o mini, selected as a general-purpose large language model providing consistent multilingual translation capabilities while maintaining computational efficiency. The translated datasets will be subsequently subjected to the same shuffling and evaluation procedures as the original-language datasets.

\subsubsection{Shuffling and Randomization}

After translation (when applicable), all datasets will undergo answer-option shuffling using a fixed random seed (42). This step will be applied uniformly across original-language datasets, translated datasets, text-only items, and multimodal items. Shuffling will be executed once per dataset version, and the resulting files will be cached and reused for all model evaluations, ensuring that all models will be tested with identical item answer ordering.

\subsubsection{Model Settings and Inference Procedure}

Inference will be performed using the GPT-5-mini model from OpenAI and the Claude-3.5-Haiku-20241022 and Claude Sonnet~4.5 models from Anthropic. All models will be accessed exclusively through the official APIs of the respective providers. No local deployment or self-hosted inference will be performed.

Because inference relies on managed cloud APIs, the underlying hardware infrastructure (e.g., CPU/GPU type or memory configuration) will not be exposed and therefore not considered in the analysis. The evaluation focuses exclusively on model outputs. No explicit temperature or sampling parameters will be specified. All models will be queried with their default decoding settings, and the maximum number of generated tokens will be constrained to the minimum required to produce the expected output format.

A strictly constrained prompting strategy will be adopted. For each instance, the model will receive a unique QuestionID, the question text, up to five answer options (A--E), and, when explicitly indicated in the dataset metadata, one or more associated medical images provided through API-compatible multimodal formats. No placeholder or synthetic images will be used. The system prompt will enforce a rigid output structure requiring a strictly formatted answer associated with the QuestionID, with no explanations or additional text. Each question will be processed independently with exactly one inference call per item. No retries, ensembling, self-consistency, or majority-voting strategies will be employed. Model outputs that did not conform to the expected format will be excluded from evaluation and not counted as incorrect predictions.

\subsubsection{Prompt Design}

A deliberately minimal and strictly constrained prompt design will be adopted to ensure comparability across models and to minimize prompt-induced variability. The prompt will be implemented using a fixed \textit{system--user message structure}, in which the system message defines strict output constraints and the user message contains the question content. A common prompt schema will be used across all datasets, providing only the information strictly necessary to perform the task, namely the question identifier, the question text, the available answer options, and, when applicable, the associated image(s).

For most language subsets, the task consists of selecting a single correct answer from a set of multiple-choice options, and the prompt enforces a single-answer output. For the French subset, where some items are defined by multiple simultaneously correct answers, the prompt structure will be minimally adapted to allow the model to return all required answer choices, while preserving the same constraints on prompt content, verbosity, and output formatting. No in-context examples, demonstrations, explanatory instructions, or reasoning scaffolds are included in any prompt variant.

\subsubsection{Specific Evaluation Rule for the French Subset}

The French subset contains items for which multiple answers may be simultaneously correct. For this subset only, a prediction will be considered correct if and only if all expected correct answers will be identified. This evaluation rule reflects the structure of the French examination items and will be applied identically to both ChatGPT and Claude models.

Questions in the France subset of EuropeMedQA may be associated with up to two distinct images, reflecting the hierarchical structure of the clinical prompt. Specifically, the dataset distinguishes between (i) a \textbf{main case image}, linked to the clinical stem of the question, and (ii) an optional \textbf{sub-question image}, associated with a more specific diagnostic or interpretative task.

The presence of these images is explicitly indicated by two separate metadata fields: \texttt{picture\_provided\_percase} (main case image) and \texttt{picture\_provided} (sub-question image). When available, the main case image is embedded as a hyperlink within the question text, whereas the sub-question image is provided through a dedicated image link column. During evaluation, multimodal inputs will be constructed as follows. The textual question will always be provided to the model. When present, the main case image will be supplied as the first visual input, followed by the sub-question image as a second visual input. Consequently, each question will be evaluated under one of four possible settings: text-only, text with main image, text with sub-question image, or text with both images.

\subsubsection{Evaluation Metrics and Data Analysis}

Model performance will be evaluated using accuracy, defined as the proportion of correctly answered questions. Each question will be treated as an independent unit of analysis, and accuracy will be computed at the question level under an implicit assumption of independence between items. Accuracy will be the primary evaluation metric because the task consists of balanced multiple-choice questions with a single discrete prediction per item, for which accuracy provides a direct and interpretable measure of model performance.

Accuracy will be computed and reported separately for the full dataset (overall accuracy), for text-only questions, and for multimodal questions (text plus image). Results will be stratified by language (original language versus English translation). No additional metrics or statistical tests beyond those implemented in the evaluation scripts will be applied.

\subsubsection{Reproducibility Considerations}

To promote reproducibility, all preprocessing and evaluation steps will be fully scripted. Fixed random seeds will be used for answer shuffling, prompting formats will be deterministic, and each item triggered exactly one model inference call. No explicit temperature or sampling parameters will be specified, and all models will be queried using their default decoding settings, avoiding the introduction of additional stochasticity at the experimental design level. All model outputs and evaluation summaries will be logged and stored, enabling independent verification of the reported results.

\section{RESIDUAL BIASES AND LIMITATIONS}

Despite the controlled design of the evaluation pipeline, some residual sources of bias and limitations remain and should be considered when interpreting the results.

First, the EuropeMedQA benchmark is derived from examination-style questions and educational materials, and therefore reflects a selection bias toward knowledge assessment and clinical reasoning under test conditions rather than the prevalence or distribution of real-world clinical cases.

Second, although answer-option shuffling will be applied to mitigate positional and class-distribution biases, the resulting uniform distribution of correct answers represents an artificial balance that improves comparability across models but does not mirror natural clinical answer frequencies.

Third, for non-English subsets, automatic translation may introduce subtle linguistic or semantic variations, which cannot be fully eliminated despite the use of a high-quality translation model and a consistent preprocessing pipeline.

Finally, inference will be performed through managed cloud APIs, limiting control over underlying hardware and low-level decoding behavior; while procedural reproducibility is ensured, strict bitwise determinism cannot be guaranteed.

These limitations do not compromise the internal validity of the comparative evaluation, but they delimit the scope of interpretation of the results to controlled, zero-shot assessment settings.

\section{ETHICAL CONSIDERATIONS}

Not applicable. All data used are publicly available from official medical examination sources, and no patient data or personally identifiable information (PII) is included. AI models will be evaluated for research purposes only, with no clinical deployment.

\section{EXPECTED OUTCOMES}

The EuropeMedQA dataset will be publicly available, promoting transparency and reproducibility in AI model evaluation. Results from model benchmarking will guide the development of more robust multimodal language models for medical applications. This dataset may contribute to filling a critical gap in multilingual and multimodal model assessment, fostering advancements in medical research.
\newpage
\printbibliography

\end{document}